\documentclass{ifacconf}
\usepackage{graphicx}  % required for figures
\usepackage{xcolor}
\usepackage{natbib}  % required for bibliography
\usepackage{amsmath}
\usepackage{amssymb}
\usepackage{mathtools}
\usepackage{textcomp}
\usepackage{bm}
\usepackage[hyphens]{url}
\usepackage{array}

\makeatletter
\renewcommand{\nsim}{\mathrel{\mathpalette\n@sim\relax}}
\newcommand{\n@sim}[2]{%
  \ooalign{%
    $\m@th#1\sim$\cr
    \hidewidth$\m@th#1\rotatebox[origin=c]{50}{$#1-$}$\hidewidth\cr
  }%
}

\let\save@mathaccent\mathaccent
\newcommand*\if@single[3]{%
     \setbox0\hbox{${\mathaccent"0362{#1}}^H$}%
     \setbox2\hbox{${\mathaccent"0362{\kern0pt#1}}^H$}%
     \ifdim\ht0=\ht2 #3\else #2\fi
}
\newcommand*\rel@kern[1]{\kern#1\dimexpr\macc@kerna}
\newcommand*\widebar[1]{\@ifnextchar^{{\wide@bar{#1}{0}}}{\wide@bar{#1}{1}}}
\newcommand*\wide@bar[2]{\if@single{#1}{\wide@bar@{#1}{#2}{1}}{\wide@bar@{#1}{#2}{2}}}
\newcommand*\wide@bar@[3]{%
     \begingroup
     \def\mathaccent##1##2{%
          \let\mathaccent\save@mathaccent
          \if#32 \let\macc@nucleus\first@char \fi
          \setbox\z@\hbox{$\macc@style{\macc@nucleus}_{}$}%
          \setbox\tw@\hbox{$\macc@style{\macc@nucleus}{}_{}$}%
          \dimen@\wd\tw@
          \advance\dimen@-\wd\z@
          \divide\dimen@ 3
          \@tempdima\wd\tw@
          \advance\@tempdima-\scriptspace
          \divide\@tempdima 10
          \advance\dimen@-\@tempdima
          \ifdim\dimen@>\z@ \dimen@0pt\fi
          \rel@kern{0.6}\kern-\dimen@
          \if#31
               \overline{\rel@kern{-0.6}\kern\dimen@\macc@nucleus\rel@kern{0.4}\kern\dimen@}%
               \advance\dimen@0.4\dimexpr\macc@kerna
               \let\final@kern#2%
               \ifdim\dimen@<\z@ \let\final@kern1\fi
               \if\final@kern1 \kern-\dimen@\fi
          \else
               \overline{\rel@kern{-0.6}\kern\dimen@#1}%
          \fi
     }%
     \macc@depth\@ne
     \let\math@bgroup\@empty \let\math@egroup\macc@set@skewchar
     \mathsurround\z@ \frozen@everymath{\mathgroup\macc@group\relax}%
     \macc@set@skewchar\relax
     \let\mathaccentV\macc@nested@a
     \if#31
     \macc@nested@a\relax111{#1}%
     \else
          \def\gobble@till@marker##1\endmarker{}%
          \futurelet\first@char\gobble@till@marker#1\endmarker
          \ifcat\noexpand\first@char A\else
               \def\first@char{}%
          \fi
          \macc@nested@a\relax111{\first@char}%
     \fi
     \endgroup
}
\makeatother

\begin{document}
\begin{frontmatter}

\title{Model Predictive Control for Cooperative Docking Between Autonomous Surface Vehicles with Disturbance Rejection\thanksref{footnoteinfo}}
\thanks[footnoteinfo]{This publication has been supported by funding from the European Union's Horizon Europe Programme under grant agreement No 101093822 (SeaClear2.0).}

% \author{Gianpietro Battocletti, Dimitris Boskos, Bart De Schutter} 
\author[First]{Gianpietro Battocletti}
\author[First]{Dimitris Boskos} 
\author[First]{Bart De Schutter}

\address[First]{Delft Center for Systems and Control, Delft University of Technology, 2628 CD Delft, The Netherlands. \\e-mails: \{g.battocletti, d.boskos, b.deschutter\}@tudelft.nl.}

\begin{abstract}
     Uncrewed Surface Vehicles (USVs) are a popular and efficient type of marine craft that find application in a large number of water-based tasks. When multiple USVs operate in the same area, they may be required to dock to each other to perform a shared task. Existing approaches for the docking between autonomous USVs generally consider one USV as a stationary target, while the second one is tasked to reach the required docking pose.
     In this work, we propose a cooperative approach for USV-USV docking, where two USVs work together to dock at an agreed location. 
     We use a centralized Model Predictive Control (MPC) approach to solve the control problem, obtaining feasible trajectories that also guarantee constraint satisfaction.
     Owing to its model-based nature, this approach allows the rejection of  disturbances, inclusive of exogenous inputs, by anticipating their effect on the USVs through the MPC prediction model. This is particularly effective in case of almost-stationary disturbances such as water currents.
     In simulations, we demonstrate how the proposed approach allows for a faster and more efficient docking with respect to existing approaches.
\end{abstract}


\end{frontmatter}
%===============================================================================

\section{Introduction}
\label{sec:introduction}
\begin{figure}[t]
     \centering
     \includegraphics[width=\columnwidth]{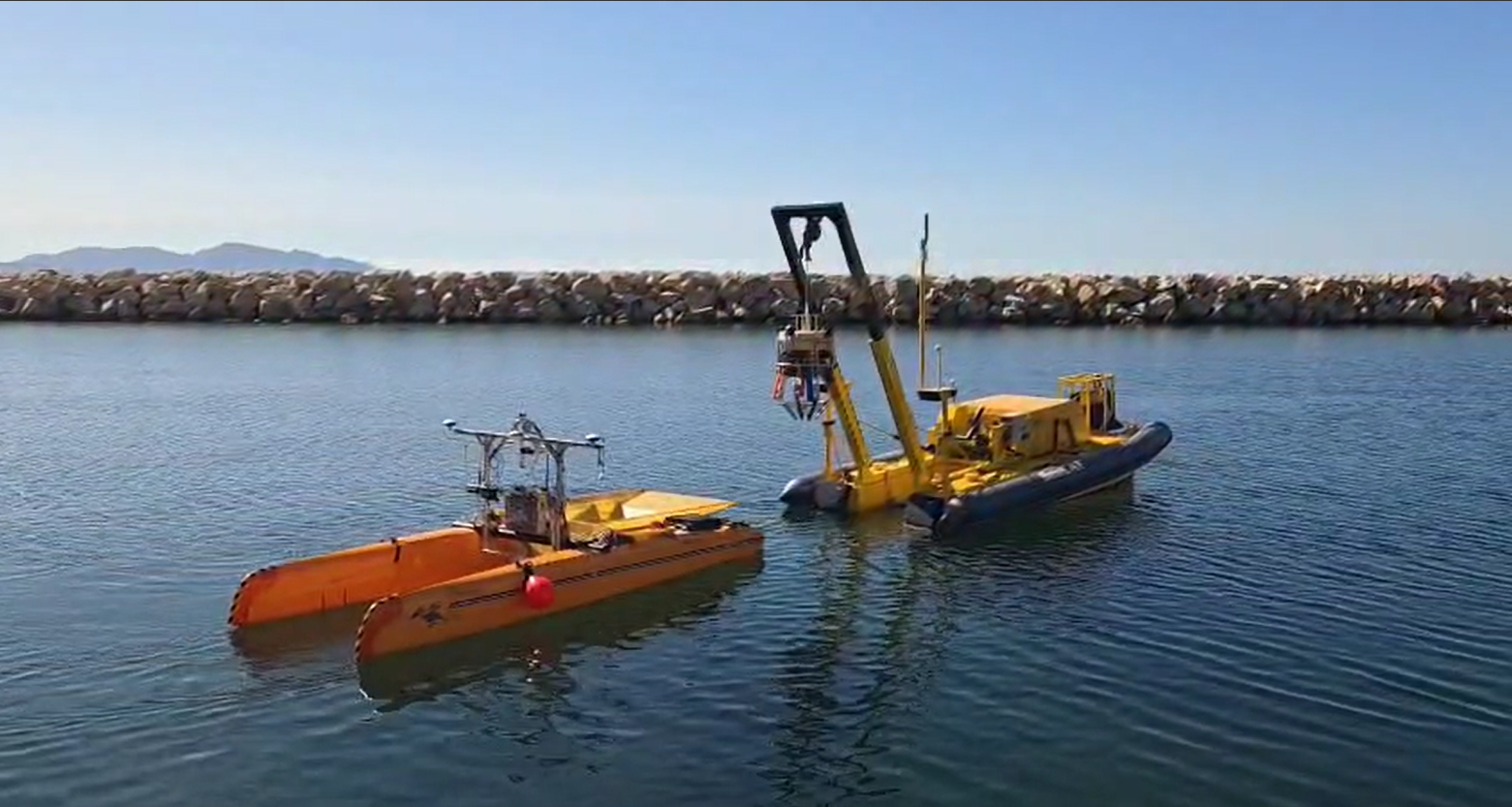}
     \caption{Example of two autonomous USVs that are about to dock so that the one with the crane (right) can deposit some material onto the other one (left).}
     \label{fig:usvs}
\end{figure}

Uncrewed Surface Vehicles (USVs) are a versatile and efficient platform for water-based operations, with applications that range from transportation to exploration, and from infrastructure inspection to litter collection \citep{bai2022review}.
USVs can be either controlled remotely or operate autonomously, with the latter option becoming increasingly more popular with the growing capabilities of planning and control algorithms \citep{liu2016unmanned}.
When multiple USVs operate in the same area, they may be specialized in specific tasks and be required to cooperate in order to complete the overall operation \citep{peng2020overview}. For instance, two USVs may be required to dock with each other to transfer a payload, as in the example shown in Figure \ref{fig:usvs}.
In a docking scenario, the USVs must approach each other and achieve a desired relative heading and position. A common approach is to require the USVs to achieve docking in a stern-to-stern or stern-to-bow configuration, with the two vessels aligned along their main axis.

Different approaches have been developed for the autonomous docking of USVs (see \citep{lexau2023automated} for a survey). The large majority of existing approaches consider a docking problem where one USV must dock to a fixed target location, e.g., a larger mothership or a ground structure \citep{gong2023mpc, hu2023design, breivik2011virtual}. In those cases, the target to dock to is considered not controllable to the purpose of docking, and the USV must reach a given pose to be able to dock to it.
This is effective when the USV must dock to a mothership with low maneuverability, as having the mothership move is more expensive than making the USV travel a longer distance, but is suboptimal when both USVs are able to move to meet at an intermediate location.
Even in most papers that consider similarly-sized USVs, such as \citep{douguet2024cooperative}, where a model-free approach based on an Artificial Potential Field (APF) is proposed, one USV is tasked to follow the other one and to dock to it.
This approach is simple, but lacks the benefits guaranteed by a cooperative approach, such as a faster and more efficient docking. 
To the best of the authors' knowledge, only a small number of works deal with cooperative docking of similarly-sized USVs.
\citet{paulos2015automated} have proposed a cooperative docking approach where multiple USVs coordinate to assemble and form a larger marine craft. However, their work focuses on the high-level coordination of the USVs, such as computing the optimal docking order, and uses a simplified graph-based search algorithm for the path planning. 
Another work that follows a similar cooperative approach is \citep{zhang2025parallel}. There, a simplified dynamics on a grid world is considered, where a set of square-shaped robots can only move left/right and up/down, and they have the task of docking to form configurations with different shapes. While this approach is effective for the omnidirectional USVs considered in that work, it is not applicable to more general marine craft such as the ones shown in Figure \ref{fig:usvs}.
Lastly, a cooperative approach based on Model Predictive Control (MPC) has been proposed by \citet{shi2023predictive}, but only for the last phase of the docking process, in which the two USVs are already close to each other with the correct heading, and they need to get in contact with each other in a precise way, e.g. to activate a coupling mechanism to keep them connected. 

In this paper, we propose an MPC-based approach for cooperative docking of two autonomous similarly-sized USVs. The proposed controller guarantees constraint satisfaction and computes dynamically feasible trajectories.
Differently from existing methods, the proposed controller focuses on cooperative trajectory planning for two autonomous marine craft that coordinate in order to reach a specific docking configuration at an agreed location.
Owing to its model-based nature, the proposed approach is able to include information about environmental disturbances due to wind, waves, and water currents, and to compensate them by predicting their effect on the USVs.
The proposed approach focuses on the trajectory planning and tracking starting from a general initial condition, until the two USVs are almost in contact, i.e., just before they touch, where a specialized controller can take over to connect the USVs \citep{shi2023predictive}.
The remainder of the paper is organized as follows. Section \ref{sec:background} introduces the problem and the dynamic model of the USVs. In Section \ref{sec:controller} the proposed controller is introduced. The simulation and evaluation of the controller is carried out in Section \ref{sec:case_study}. Finally, Section \ref{sec:conclusions} concludes the paper.

\section{Background}
\label{sec:background}
\subsection{Notation}
\label{sec:notation}
We define reference frames using the common notation $\{b\} = (O_b, \hat{x}_b, \hat{y}_b)$, where $O_{b}$ is the origin of the frame, and $\hat{x}_b$, $\hat{y}_b$ are orthogonal unit vectors \citep{fossen2011handbook}.
When needed, we use superscripts in round brackets to indicate the reference frame in which a vector is expressed, e.g., $v^{(b)}$.
To transform vectors between different reference frames we use 2D rotation matrices $R \in \mathrm{SO}(2)$ from the 2D special orthogonal group. 
We indicate with a superscript the target reference frame and with a subscript the reference frame of the original vector, e.g., $v^{(e)} = R^{(e)}_{(b)} v^{(b)}$.
For convenience, we use the shorthand expression for the sine and cosine, function, i.e., $s_\alpha \coloneqq \sin(\alpha)$.
We use superscripts in square brackets to index the entries of vectors, e.g., $v^{[1]} = a$ for $v = [a, b]^\top$.

\subsection{Dynamic Model of the USVs}
\label{sec:model}
\begin{figure}[t]
     \centering
     \includegraphics{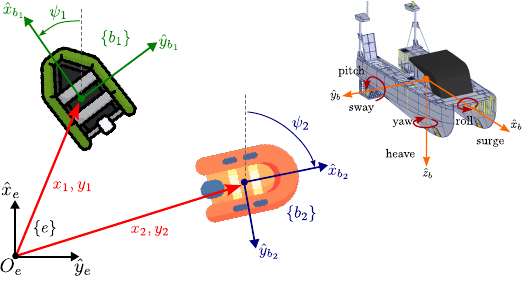}
     \caption{Representation of the reference frames used to define the dynamic models of the two USVs. The top-right corner shows a 3D model of a USV with its 6 degrees of freedom.}
     \label{fig:reference_frames}
\end{figure}

The most common approach when dealing with station keeping and low-speed maneuvering of USVs is to consider a 2D point-mass nonlinear dynamic model, in which only the surge, sway, and yaw dynamics are modeled \citep{fossen2011handbook}, as shown in Figure \ref{fig:reference_frames}.
In such a model, the state of a USV is given by the vector $q = [\eta^\top, \nu^\top]^\top \in \mathbb{R}^6$, where $\eta = [x, y, \psi]^\top$ is the position and orientation vector expressed in the inertial reference frame $\{e\}$, and $\nu = [u, v, \omega]^\top$ is the velocity vector expressed in the body reference frame $\{b\}$.
The dynamic model is then
\begin{equation}
     \label{eq:nonlinear_model}
     \begin{aligned}
          & M \dot{\nu} + C(\nu)\nu + D_\mathrm{L}\nu + D_\mathrm{NL}(\nu)\nu = b + \tau\\
          &\dot{\eta} = \widetilde{R}(\psi)\nu,
     \end{aligned}
\end{equation}
where $M$ is the inertia matrix, $C$ is the Coriolis matrix, $D_\mathrm{L}$ and $D_\mathrm{NL}$ are the linear and nonlinear damping matrices, respectively, and $\widetilde{R}(\psi) \coloneqq \mathrm{diag}(R(\psi), 1)$ is the transformation matrix to transform the coordinates of the USV between the local and global reference frames (see Figure \ref{fig:reference_frames}), with $R$ indicating a 2D rotation matrix.
The inputs of the system are the force vector $\tau$, which is expressed as a force vector acting on the center of mass of the USV, and the bias vector $b$, which collects the hydrodynamic forces due to wind, waves, and ocean currents \citep{fossen2011handbook}.
In this work, the vector $b$ is considered as stationary in the inertial reference frame $\{e\}$, i.e., $\dot{b}^{(e)}\approx 0$, as the changes in direction and magnitude of wind, waves, and water currents over the duration of the docking operation are in general negligible \citep{fossen2009kalman}.
Additionally, for the purpose of this paper, the high frequency effects of waves, which have zero mean, are neglected \citep{fossen2009kalman}.

The nonlinearities in \eqref{eq:nonlinear_model} are due to (i) the Coriolis matrix $C(\nu)$, (ii) the nonlinear damping term $D_\text{NL}(\nu)$, and (iii) the rotation matrix $R(\psi)$. The matrices $C(\nu)$ and $D_\text{NL}(\nu)$ can be neglected under the assumption of low-speed maneuvering (for which the threshold is usually set as 6 knots, i.e., 2 m/s) \citep{fossen2011handbook}. The resulting dynamic model becomes
\begin{equation}
     \label{eq:dynamic-linear-model}
     \begin{aligned}
          & M \dot{\nu} + D_L\nu = b + \tau \\
	     &\dot{\eta} = \widetilde{R}(\psi) \nu.
     \end{aligned}
\end{equation}

In the models \eqref{eq:nonlinear_model}, \eqref{eq:dynamic-linear-model}, the force vector $\tau$ depends on the control vector $u$, which assigns a force to each of the actuators of the USV.
The relation between $u$ and $\tau$ depends on the type of the thrusters and their configuration, and is, in general, specific to each USV. Here we present two illustrative thruster configurations inspired by the USVs used in the SeaClear2.0\footnote{\url{https://www.seaclear2.eu/}} project, but it is important to note that the proposed approach is general and also allows different configurations.
In this work we consider two USVs: $\mathrm{USV}_1$, which is equipped with four fixed thrusters, two at the stern and two tilted sideways on the bow (represented in Figure \ref{fig:reference_frames} by the green USV, and also by the 3D model in the top-right corner of the figure), and $\mathrm{USV}_2$, which is equipped with two azimuth thrusters that can rotate around the vertical axis (represented by the orange USV in Figure \ref{fig:reference_frames}).

For $\mathrm{USV}_1$ the relation between the control vector $u_1$ and the force vector $\tau_1$ is 
\begin{equation}
     \label{eq:tau1_u1}
     \tau_1 = \begin{bmatrix}
		1 & 1 & s_\alpha &  s_\alpha \\
		0 & 0 & c_\alpha & -c_\alpha \\
		w_1 & -w_1 & d_1 c_\alpha + w_1 s_\alpha & -d_1 c_\alpha - w_1 s_\alpha
	\end{bmatrix} u_1.
\end{equation}
In \eqref{eq:tau1_u1}, $w_1$, and $d_1$ are geometric parameters related to the USV width and length, while $\alpha$ is the angle at which the bow thrusters are tilted with respect to the $\hat{y}$ axis.
Furthermore, $u_1\in\mathbb{R}^4$ is the vector of the control inputs to the four thrusters, with $u_1^{[1]}, u_1^{[2]}$ indicating the forces of the stern thrusters, and $u_1^{[3]}, u_1^{[4]}$ the forces of the bow thrusters.

$\mathrm{USV}_2$ is equipped with two azimuth thrusters, and therefore the relationship between $\tau_2$ and $u_2$ is nonlinear:
\begin{equation}
     \tau_2 = \begin{bmatrix*}
          u_2^{[1]} c_{u_2^{[3]}} + u_2^{[2]} c_{u_2^{[4]}} \\
          -u_2^{[1]} s_{u_2^{[3]}} - u_2^{[2]} s_{u_2^{[4]}} \\
          w_2(u_2^{[1]} c_{u_2^{[3]}} - u_2^{[2]} c_{u_2^{[4]}}) + d_2( u_2^{[1]} s_{u_2^{[3]}} - u_2^{[2]} s_{u_2^{[4]}})
     \end{bmatrix*}.
\end{equation}
Here, $u_2^{[1]}$ and $u_2^{[2]}$ represent the actuator forces, $u_2^{[3]}$ and $u_2^{[4]}$ represent the rotation angles of the azimuth thrusters, and $w_2$ and $d_2$ are geometric parameters related to the dimensions of the USV. 

\subsection{Problem Setting}
\label{sec:problem}
We consider two autonomous USVs that are tasked to dock with each other (see Figure \ref{fig:usvs}). For a successful docking, the USVs must align their sterns and approach each other until they get in contact. 
In this work, we only focus on the coordinated trajectory planning of the two USVs up to the moment at which they get in contact. Once this has been achieved, we assume the availability of a dedicated controller or coupling mechanism that takes over to keep the USVs docked.
The USVs start from initial locations $q_{1, \mathrm{init}}$, $q_{2, \mathrm{init}}$ and must achieve a final configuration such that (i) $\psi_1 = \psi_2 + \pi$, i.e., the sterns of the USVs face each other, (ii) the distance between the centers of mass of the USVs along the $\hat{x}_{b_1}$ axis is equal to $d_\mathrm{min} = (d_1 + d_2)/2$, which means that the hulls are in contact, and (iii) the distance between the USVs along the $\hat{y}_{b_1}$ axis is equal to $0$.

\section{Proposed cooperative docking approach}
\label{sec:controller}
\begin{figure}[t]
     \centering
     \includegraphics[]{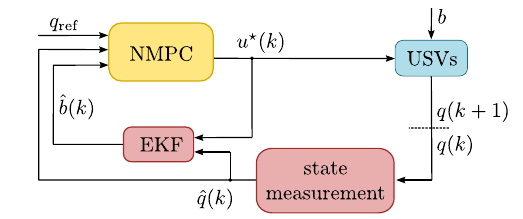}
     \caption{Schematic representation of the controller architecture.}
     \label{fig:controller_architecture}
\end{figure}

In this section, we outline the different components of the proposed controller to achieve the cooperative docking. The controller architecture is visualized in Figure \ref{fig:controller_architecture}. 

\subsection{State Measurement and Disturbance Estimation}
\label{sec:disturbances-estimation}
We assume the availability of dedicated sensors and observers to measure and/or estimate the state of the USVs. The result of this process is used as the initial condition of the MPC optimization problem (cf. Figure \ref{fig:controller_architecture}). Additionally, the information about the state of the USVs is used to estimate the disturbances and exogenous inputs acting on the USVs.
In maneuvering theory, the estimation of the disturbances affecting the motion of a marine craft has been studied extensively (see \citep{fossen2011handbook,fossen2009kalman, wirtensohn2016disturbance}). Several types of estimation techniques have been proposed, both to estimate the low-frequency effects of exogenous inputs such as water current, wind, and waves, and to estimate the effect of higher frequency components of waves through techniques such as wave filtering \citep{fossen2009kalman}. 
In this work, we assume the availability of a model-based observer to estimate the lumped low-frequency exogenous input $b$. 
In particular, we consider an Extended Kalman Filter (EKF) operating with an augmented dynamic model that includes $b$ as part of the state \citep{fossen2009kalman}. 
The EKF estimate, which we indicate as $\hat{b}$, is then used by the MPC prediction model to predict the effect of $b$ on the USVs.
In principle, the EKF can be replaced with other types of observers \citep{wirtensohn2016disturbance, allotta2017sea}, including multi-agent observers, which allow to exploit the availability of multiple sensors on multiple USVs \citep{wu2019cooperative}.

\subsection{NMPC Controller}
\label{sec:NMPC}
We formulate the control problem as a nonlinear optimization problem to be solved in a receding-horizon fashion through an MPC controller. Given the low number of agents involved and the relatively slow dynamics of the system, we consider a centralized controller. 
The controller can be run on board on one of the two USVs or on a central computation unit (either on a third vessel or on land). In both cases, we assume the availability of a reliable communication channel through which the USVs can exchange all the necessary information, such as the state measurement and the control action to apply (i.e., the solution to the MPC optimization problem).
The resulting MPC optimization problem is a nonlinear program (NLP), which can be solved using numerical optimization algorithms. Gradient-based methods such as multi-start Sequential Quadratic Programming (SQP) are usually preferred, but gradient-free approaches such as a genetic algorithm could be applied as well.

We consider a discrete-time version of the dynamics introduced in Section \ref{sec:model}. We use the variable $k$ as a discrete-time counter for time steps of $\Delta t$ time units. 
We consider the joint state $q(k) = [q_1^\top(k \Delta t), q_2^\top (k \Delta t)]^\top \in \mathbb{R}^{12}$, the joint control vector $u(k) = [u_1^\top (k \Delta t), u_2^\top (k \Delta t)]^\top \in \mathbb{R}^8$, and the joint discrete-time dynamics $q(k+1) = f(q(k), u(k), b)$.
The dynamics $f$ are computed as a discrete-time approximation of the continuous one \eqref{eq:dynamic-linear-model} using the forward Euler discretization.
The centralized NMPC controller with prediction horizon $N > 1$ is then defined by the following NLP:
\begin{subequations}
	\label{eq:NMPC}
	\begin{align}
		J\big(q(k), &u(k-1), \hat{b}(k), q_\mathrm{ref}) = \nonumber \\*
		&\hspace{-12pt}\min_{\textbf{q}(k), \textbf{u}(k)} \sum_{h=1}^{N} J_\mathrm{tracking}\Big(q(h|k), q_\mathrm{ref}\Big) \nonumber \\
		&\quad\quad\quad + \sum_{h=1}^{N-1} J_\mathrm{control}\Big(u(h|k)\Big) \\
		&\text{s.t.} \!\quad q(0|k) = q(k) \label{eq:IC-x}\\
		&\qquad u(-1|k) = u(k-1) \label{eq:IC-u}\\
		&\qquad\text{for} \quad h = 0,\dots,N-1: \nonumber \\
		\vspace{0.5cm}
		&\qquad\quad q(h+1|k) = f\big(q(h|k), u(h|k), \hat{b}(k)\big) \label{eq:dynamics-mpc}\\
          &\qquad\quad F_\mathrm{min} \leq u(h|k) \leq F_\mathrm{max}  \label{eq:input_constraint} \\
          &\qquad\text{for} \quad h = 0,\dots,N: \nonumber \\
          &\qquad\quad\left\Vert\begin{bmatrix}
                         q^{[1]}(h|k) \\ 
                         q^{[2]}(h|k) \\ 
                    \end{bmatrix} 
                    - \begin{bmatrix}
                         q^{[7]}(h|k) \\
                         q^{[8]}(h|k) \\ 
                    \end{bmatrix}\right\Vert_2 \geq d_\mathrm{min} \label{eq:collision-avoidance}\\
          &\qquad\quad\left\Vert [q^{[4]}(h|k), q^{[5]}(h|k)]^\top \right\Vert_2 \leq v_{\mathrm{max}, 1}, \label{eq:constraint-speed-1} \\
          &\qquad\quad\left\Vert [q^{[10]}(h|k), q^{[11]}(h|k)]^\top \right\Vert_2 \leq v_{\mathrm{max}, 2} \label{eq:constraint-speed-2}\\
          &\qquad\text{for} \quad h = 0,\dots,N-1: \nonumber \\
		&\qquad\quad |u(h|k) - u(h-1|k)| \leq \Delta F_\mathrm{max} \Delta t, \label{eq:input_constraint_rate}
	\end{align}
\end{subequations}
where $q(h|k)$ and $u(h|k)$ are the predicted states and inputs, respectively, $h$ steps into the prediction window period of the MPC optimization problem at time step $k$.
Bold variables gather a variable over the prediction horizon, e.g., \begin{equation}
	\begin{aligned}
		\textbf{q}(k) = \big[q^\top(0|k), \dots, q^\top(N|k)\big]^\top.
	\end{aligned}
\end{equation}

The inputs to the MPC problem are the joint state $q(k)$, the control input applied at the previous time step $u(k-1)$ (which is used to enforce the input constraint \eqref{eq:input_constraint_rate}), the estimated exogenous input vector $\hat{b}(k)$, and the state reference vector $q_\mathrm{ref}$, defined as
\begin{equation}
     q_\mathrm{ref} = [x_{\mathrm{ref}, 1}, y_{\mathrm{ref}, 1}, \psi_{\mathrm{ref}, 1}, x_{\mathrm{ref}, 2}, y_{\mathrm{ref}, 2}, \psi_{\mathrm{ref}, 2}]^\top \in \mathbb{R}^6.
\end{equation}
The first three terms of $q_\mathrm{ref}$ represent the target pose for USV$_1$, while the last three represent the target pose for USV$_2$. For simplicity, we do not set a reference for the velocity components of $q$, relying instead on the MPC controller to bring them to zero once the target positions and headings are reached. 
In order to ensure a docking that satisfies the specifications outlined in Section \ref{sec:problem}, $q_\mathrm{ref}$ must satisfy the conditions
\begin{equation}
     \left\Vert\begin{bmatrix}
          x_{\mathrm{ref}, 1} \\ 
          y_{\mathrm{ref}, 1} \\ 
     \end{bmatrix} 
     - \begin{bmatrix}
          x_{\mathrm{ref}, 2} \\
          y_{\mathrm{ref}, 2} \\ 
     \end{bmatrix}\right\Vert_2 \geq d_\mathrm{min},
\end{equation}
and
\begin{equation}
     \psi_{\mathrm{ref}, 1} = \psi_{\mathrm{ref}, 2} + \pi.
\end{equation}
In the following, we assume $q_\mathrm{ref}$ to be provided externally, e.g. by another module in the control system. In the case study in Section \ref{sec:case_study} we will compare different options to select $q_\mathrm{ref}$.

The cost function is composed of two terms, namely:
\begin{enumerate}
     \item The tracking cost $J_\mathrm{tracking}$, defined as
     \begin{align}
          J_\mathrm{distance}(q, q_\mathrm{ref}) = \xi_1^\top Q_1 \xi_1 + \xi_2^\top Q_2 \xi_2.\label{eq:cost_tracking}
     \end{align}
     This component of the cost function has the purpose of encouraging the USVs to reach the position and heading set by $q_\mathrm{ref}$.
     The vectors $\xi_1$ and $\xi_2$ represent the position and heading error of the two USVs with respect to $q_\mathrm{ref}$, and are defined as 
     \begin{align}
          \label{eq:xi}
          \xi_1 = \begin{bmatrix}
               q^{[1]} - q_\mathrm{ref}^{[1]}\\
               q^{[2]} - q_\mathrm{ref}^{[2]}\\
               q^{[3]} - q_\mathrm{ref}^{[3]}
          \end{bmatrix}, \quad
          \xi_2 = \begin{bmatrix}
               q^{[7]} - q_\mathrm{ref}^{[4]}\\
               q^{[8]} - q_\mathrm{ref}^{[5]}\\
               q^{[9]} - q_\mathrm{ref}^{[6]}
          \end{bmatrix}.
     \end{align}
     The matrices $Q_1, Q_2 \in \mathbb{R}^{3\times3}, Q_1, Q_2 \succ 0$ are diagonal weight matrices to balance the relative importance of position and heading tracking. 
     \item The term $J_\mathrm{control}$, which aims at keeping the control effort as low as possible, defined as
     \begin{equation}
          \label{eq:cost_input}
          J_\mathrm{control}(u) = u^\top Q_u u,
     \end{equation}
     where $Q_u \in \mathbb{R}^{8 \times 8}, Q_u \succ 0$ is a weight matrix balancing the relative importance of the different control inputs.
\end{enumerate}

The constraints \eqref{eq:IC-x} and \eqref{eq:IC-u} set the initial conditions in the MPC optimization problem. The dynamics of the USVs are enforced by constraint \eqref{eq:dynamics-mpc}, while \eqref{eq:input_constraint} and \eqref{eq:input_constraint_rate} represent constraints on the inputs and on the input rate, with $F_\mathrm{min}, F_\mathrm{max} \in \mathbb{R}^8$ being the bounds on the inputs, and $\Delta F_\mathrm{max} \in \mathbb{R}^8$ being the bound on the input rate.
The state constraints are defined in \eqref{eq:collision-avoidance}, \eqref{eq:constraint-speed-1}, and \eqref{eq:constraint-speed-2}, where \eqref{eq:constraint-speed-1} and \eqref{eq:constraint-speed-2} represent the velocity constraints of the two USVs, while \eqref{eq:collision-avoidance} is a collision avoidance constraint, imposing a minimum distance $d_\mathrm{min}$ between the points $O_{b_1}$ and $O_{b_2}$.
In principle, the collision avoidance constraint should take into account the shape of the USVs, which can be represented as rectangles centered in $O_{b_1}$ and $O_{b_2}$, and it should avoid intersections between these rectangles \citep{lavalle2006planning}.
In our approach, we use the simplified version \eqref{eq:collision-avoidance} of this constraint, in which we consider the USVs as disk-shaped, in order to make it more easily tractable in the MPC problem.

At every time step $k$, the MPC optimization problem \eqref{eq:NMPC} is solved, returning an optimal control sequence. The first component of the optimal control sequence, which we denote as $u^\star(k) = u(0|k)$, is then applied to the system, while the rest is discarded. 
The problem is then solved again at the next time step, in a receding-horizon fashion. 

\section{Case study}
\label{sec:case_study}
We provide an illustrative example of the proposed cooperative docking controller.
The two USVs considered in the case study are those shown in Figure \ref{fig:usvs}. 
The values of their model parameters are listed in Table \ref{tab:model-values}.
Also in this case, the values of the parameters (including the input constraints reported below) have been inspired by the USVs used in the SeaClear2.0 project.

\begin{table}[h]
     \centering
     \caption{Values of the USVs model parameters.}
     \label{tab:model-values}
     \begin{tabular}{ccc}
          \hline
          \textbf{Parameter} & $\bm{\mathrm{USV}_1}$ & $\bm{\mathrm{USV}_2}$ \\
          \hline
          $M$ &
          $\begin{bmatrix} 1426& 0 & 0 \\ 0 & 3250 & 130 \\ 0& 130 & 7619 \end{bmatrix}$ & 
          $\begin{bmatrix} 774& 0 & 0 \\ 0 & 1625 & 0 \\ 0& 0 & 3810 \end{bmatrix}$\\
          $D_\mathrm{L}$ &
          $\begin{bmatrix} 343& 0 & 0 \\ 0 & 825 & 33 \\ 0& 33 & 1890 \end{bmatrix}$ & 
          $\begin{bmatrix} 704& 0 & 0 \\ 0 & 412 & 0 \\ 0& 0 & 945 \end{bmatrix}$ \\
          $d_i$ & 2.1 & 1.5 \\
          $w_i$ & 0.8 & 0.85 \\
          $\alpha$ & 15 & -- \\
          \hline
     \end{tabular}
\end{table}

The velocity bounds are set to $v_{\mathrm{max}, 1} = 3.3$ m/s for USV$_1$, $v_{\mathrm{max}, 2} = 3.0$ m/s for USV$_2$, and the input constraints to 
\begin{align*}
     &F_\mathrm{max} = [1000, 1000, 170, 170, 100, 100, 6.28, 6.28],\\
     &F_\mathrm{min} = [-800, -80, -170, -170, -100, -100, -6.28, -6.28],\\
     &\Delta F_\mathrm{max} = [500, 500, 85, 85, 50, 50, 0.5, 0.5],\\
     &\Delta F_\mathrm{min} = [-400, -400,   -85, -85, -50, -50, -0.5, -0.5].
\end{align*}
The controller parameters have been set as follows. The time step is set to $\Delta t = 0.5$ s, the horizon to $N=50$, and the weight matrices to $Q_1 = Q_2 = I_{3\times3}\cdot10^3$ (with $I_{3\times3}$ indicating the $3\times3$ identity matrix). The matrix $R$ is set to $R = \mathrm{diag}(10^{-2}, 10^{-2}, 10^{-3}, 10^{-3}, 10^{-2}, 10^{-2}, 10^{-4}, 10^{-4})$.

\subsection{Comparison Controllers}
In the numerical experiments we consider the following controllers:

\textit{1) Cooperative docking:} This controller is based on the proposed cooperative docking approach, which solves the MPC optimization problem \eqref{eq:NMPC}.
We test this controller under two operating condition. First, we evaluate the controller in nominal conditions, i.e., as described in Section \ref{sec:controller}. Second, we disable the disturbance estimation module, i.e., we set $\hat{b} = [0, 0, 0]^\top$, in order to assess the benefit of knowing $b$ during the docking. 

\textit{2) Docking to stationary USV (baseline controller):} This controller maintains the pose of USV$_1$ constant, while the USV$_2$ is tasked to move to complete the docking. The MPC optimization problem is identical to \eqref{eq:NMPC}, except for the fact that the first three elements of the reference vector $q_\mathrm{ref}$, i.e., $x_{\mathrm{ref}, 1}$, $y_{\mathrm{ref}, 1}$, and $\psi_{\mathrm{ref}, 1}$, coincide with the initial location of USV$_1$, which is therefore tasked to maintain its position in a station keeping fashion \citep{fossen2011handbook}. 
In this way, USV$_1$ is able to counteract the exogenous inputs $b$, and can be considered in the collision avoidance constraint.
This controller corresponds to the approach followed by most of the existing works on 
autonomous docking of USVs, e.g., \citep{breivik2011virtual}, and is therefore considered as a baseline against which the proposed approach and its variations are compared.

The following metrics are measured during the simulations:
\begin{itemize}
     \item The relative performance gain $\Delta\widetilde{J}$ with respect to the baseline controller, defined as
     \begin{equation}
          \Delta\widetilde{J} = \frac{\widetilde{J}_\mathrm{baseline} - \widetilde{J}_\mathrm{controller}}{\widetilde{J}_\mathrm{baseline}}.
     \end{equation}
     The performance $\widetilde{J}$ is computed using the following discrete-time cost function:
     \begin{equation}
          \label{eq:comparison_cost}
          \widetilde{J} = \sum_{k=0}^{\tilde{N}} J_\mathrm{tracking}(q(k), q_\mathrm{ref}) + J_\mathrm{control}(u^\star(k)),
     \end{equation}
     where $\widetilde{N} \in \mathbb{N}$ is the duration of the simulation expressed as a number of discrete time steps, and the two cost terms are the same as those used in \eqref{eq:NMPC}, defined in \eqref{eq:cost_tracking} and \eqref{eq:cost_input}.
     \item The docking time $T = K\Delta t$ after which the USVs are always closer to their reference than a threshold distance that we set to 0.5 m, and with a heading error less than 10\textdegree. 
     \item The lengths $l_1$, $l_2$ of the trajectories traveled by the two USVs.
\end{itemize}

The simulations have been implemented in Python 3.11.11 by adapting the simulation environment of the well-established MSS simulator \citep{fossen2021python}. 
The simulations have been run on a laptop with an 11th Gen Intel processor with four i7 cores (3.00 GHz) and 16 Gb of RAM.
The MPC optimization problems are solved with Ipopt \citep{wachter2006implementation}.
Source code is available at \url{https://github.com/gbattocletti/usvs-docking}.

\subsection{Numerical Experiments}
\begin{figure*}[t]
     \centering
     \includegraphics[]{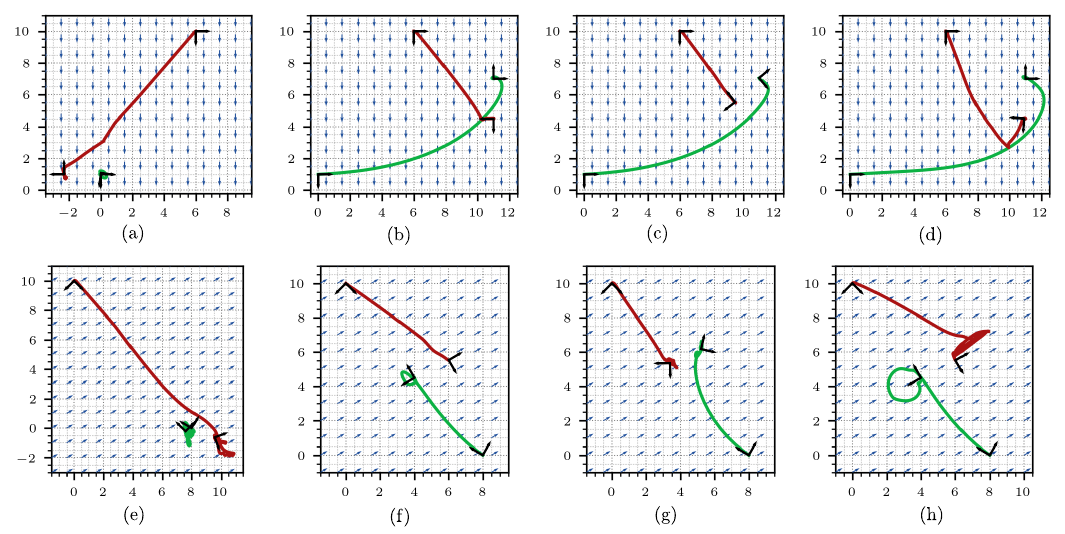}
     \vspace{-0.4cm}
     \caption{Comparison of the docking controller in two test scenarios. (a)-(d) The evaluation of the controllers in the scenario \#1. (e)-(h) The evaluation of the controllers in the scenario \#2. In the plots, the blue arrows indicate the water current, corresponding to the disturbance $b$. The green and red curves represent the paths traveled by USV$_1$ and USV$_2$, respectively. A reference frame indicating the heading of the USV is depicted in the initial and final state of each USV. Note that the reference frames are represented using the NED convention \citep{fossen2011handbook}, as shown also in Figure \ref{fig:reference_frames}.}
     \label{fig:evaluation}
\end{figure*}
We simulate the controllers in two different scenarios, with the USVs starting from different initial conditions, and with different exogenous disturbances. 
In the first scenario (indicated as \#1), a water current having speed 0.15 m/s and heading $\pi$ is considered, while in the second scenario (indicated as \#2) the water current is set to a speed of 0.25 m/s and heading $\pi/3$.
In each scenario, we perform 4 evaluations in total.
First, we evaluate the baseline controller.
Second, we select two different docking configurations (i.e., with different target locations and headings for the two USVs), which we indicate as $q_{\mathrm{ref}, 1}$, $q_{\mathrm{ref}, 2}$, and evaluate the proposed cooperative approach in each of them to investigate the effect of the choice of the docking configuration on the controller performance. 
In particular, we select docking configuration that have similar target locations, but different target headings.
In $q_{\mathrm{ref}, 1}$, the docking heading is close to the heading of the water current, while in $q_{\mathrm{ref}, 2}$ it is selected arbitrarily.
Lastly, as previously mentioned, we also evaluate the cooperative controller with $\hat{b}=0$. In this case, we select the target docking configuration to be $q_{\mathrm{ref}, 1}$, i.e., the same of one of the two evaluations of the cooperative controller in nominal conditions.
For clarity, it is worth highlighting that $q_{\mathrm{ref}, 1}$ and $q_{\mathrm{ref}, 2}$ are different between the two experiments, to account for the different initial conditions.
The trajectories obtained from the simulation of the controllers in the two scenarios are shown in Figure \ref{fig:evaluation}.
Note that the trajectories end at a distance of $d_\mathrm{min}$ m (to account for the size of the USVs), and with an offset of $\pi$ between the headings, as specified in Section \ref{sec:problem}.
An animated version of the simulations is available in the code repository at \url{https://github.com/gbattocletti/usvs-docking/scripts/results}.

The values of the metrics evaluated in the two scenarios for each controller are reported in Table \ref{tab:results}. 
\begin{table}[h]
     \centering
     \caption{Evaluation metrics computed in the two test scenarios.}
     \label{tab:results}
     \begin{tabular}{wc{0.1cm}wc{0.15cm}ccccc}
          \hline
          \textbf{\#} & \textbf{Plot} & \textbf{Controller} & $\Delta \widetilde{J}$ [\%]& $l_1$ [m] & $l_2$ [m] & $T$ [s] \\
          \hline
          1 & (a) & baseline & 0.00 & 1.64 & 14.81 & 101.5 \\
          1 & (b) & coop. $q_{\mathrm{ref}, 1}$ & 41.4 & 15.05 & 8.42 & 32.0 \\
          1 & (c) & coop. $q_{\mathrm{ref}, 2}$ & 56.3 & 14.34 & 5.86 & 45.5 \\
          1 & (d) & coop. $\hat{b}=0$ & 35.0 & 16.38 & 11.12 & 91.0 \\
          \hline
          2 & (e) & baseline & 0.00 & 16.86 & 21.82 & -- \\
          2 & (f) & coop. $q_{\mathrm{r ef}, 1}$ & 85.3 & 8.40 & 7.52 & 25.5 \\
          2 & (g) & coop. $q_{\mathrm{ref}, 2}$ & 87.2 & 9.80 & 8.13 & 80.5 \\
          2 & (h) & coop. $\hat{b}=0$ & 66.9 & 12.28 & 16.29 &  161.5 \\
          \hline
     \end{tabular}
\end{table}

By comparing the plots corresponding to the evaluation of the controllers, the benefits of the proposed cooperative approach stand out. 
In both scenarios, the proposed controller manages to reach the desired docking configuration faster than the baseline, which in scenario \#2 does not reach the desired configuration within the required tolerance.
This is likely due to the fact that the water current is badly aligned with the initial heading of USV$_1$, which in the case of the baseline controller coincides with the docking heading, creating a sideways force on the USVs that is difficult to compensate.
This suggests that aligning the $\hat{x}$ axis of the USVs with the disturbances vector $b$, when it is known, can help the USVs to reach and maintain the docking configuration more efficiently. To verify this hypothesis, the docking heading of $q_{\mathrm{ref},1}$ is set to match the disturbance heading, both in scenario \#1 and in scenario \#2. In $q_{\mathrm{ref},2}$, instead, an arbitrary heading is selected.
The effect of this choice can be observed by comparing the plots in Figures 4(b) and 4(c), as well as 4(f) and 4(g). It can be observed how, in Figure 4(g), the USVs end the docking sideways with respect to the water current, requiring more corrections to reach the docking configuration, and resulting in a longer docking time (see also the corresponding rows in Table \ref{tab:results}).
The superior performances of the proposed approach are confirmed by the numerical results in Table \ref{tab:results}, where the proposed approach presents a significant performance gain with respect to the baseline in terms of the cost function \eqref{eq:comparison_cost}. 
In addition, the benefits of estimating $\hat{b}$ are shown by the difference between the plots 4(b) and 4(d). Plot 4(b) corresponds to the proposed approach in nominal conditions, while 4(d) corresponds to the case in which the USVs aim to reach the same configuration as in 4(b), but without knowledge of the disturbance (i.e., $\hat{b}=0$). From the comparison of the two plots, it is clear that in plot 4(d) the USVs have to travel a longer distance, and initially overshoot the target location (cf. red curve). The same trend is observable even more clearly in plots 4(f) and 4(h). In the latter, the USVs have to make significant effort to reach the docking location, requiring almost 3 minutes to reach it, against the 25 s required by the proposed approach in nominal conditions.
These comparisons show the benefits of having an estimate of the exogenous input force $b$, and how the proposed controller can make use of this information to compute more efficient trajectories.

\section{Conclusions}
\label{sec:conclusions}
We have proposed a cooperative docking controller for the docking between two similarly sized USVs at a desired location.
We proposed the use of a centralized MPC controller for the cooperative docking, obtaining feasible trajectories that also guarantee constraint satisfaction.
We have shown the advantages of the proposed controller over existing approaches that consider one USV as stationary.
Moreover, we have shown how the proposed approach can make use of an estimation of the exogenous inputs acting on the USVs to counteract them by predicting their effect. We have also shown how the knowledge of the heading of the exogenous inputs can be exploited to select the docking heading, in order to obtain a faster and more efficient docking.

The main future work direction regards the improvement of the computation of the reference state that sets the docking location and heading, e.g., by optimizing it for maximum performance. Alternatively, a cost function can be designed to enable the controller to autonomously select the best docking location. 
Future work will also look at making the controller more robust to measurement noise and disturbances.
Lastly, the proposed controller should be tested in a more realistic simulator and in the real-world to assess its capabilities and robustness in real-life scenarios.

\bibliography{references}

\end{document}